# Language, Emotions, and Cultures: Emotional Sapir-Whorf Hypothesis

Leonid Perlovsky
Harvard University and the AFRL, E-mail: leonid@seas.harvard.edu

Abstract - An emotional version of Sapir-Whorf hypothesis suggests that differences in language emotionalities influence differences among cultures no less than conceptual differences. Conceptual contents of languages and cultures to significant extent are determined by words and their semantic differences; these could be borrowed among languages and exchanged among cultures. Emotional differences, as suggested in the paper, are related to grammar and mostly cannot be borrowed. Conceptual and emotional mechanisms of languages are considered here along with their functions in the mind and cultural evolution. In language evolution from primordial undifferentiated animal cries conceptual contents increase, emotional reduced. This leads to a fundamental contradiction in human mind. Reduced emotionality of vocalization is necessary for language evolution. But "too low" emotionality makes language "irrelevant to life," disconnected from sensory-motor experience. Neural mechanisms of these processes are suggested as well as their mathematical models: the knowledge instinct, the language instinct, the dual model connecting language and cognition, dynamic logic, neural modeling fields. Mathematical results are related to cognitive science, linguistics, and psychology. Experimental evidence and theoretical arguments are discussed. Approximate equations for evolution of human minds and cultures are obtained. The knowledge instinct operating in the mind hierarchy leads to mechanisms of differentiation and synthesis determining ontological development and cultural evolution. The mathematical model identifies three types of cultures: "conceptual" pragmatic cultures, in which emotionality of language is reduced and differentiation overtakes synthesis resulting in fast evolution at the price of uncertainty of values, self doubts, and internal crises; "traditional-emotional" cultures where differentiation lags behind synthesis, resulting in cultural stability at the price of stagnation; and "multi-cultural" societies combining fast cultural evolution and stability. Unsolved problems and future theoretical and experimental directions are discussed.

## 1. Emotional Sapir-Whorf Hypothesis

Benjamin Whorf (Whorf, 1956) and Edward Sapir (Sapir, 1985) in a series of publications in the 1930s researched an idea that the way people think is influenced by the language they speak. Although there was a long predating linguistic and philosophical tradition, which emphasized influence of language on cognition (Bhartrihari, IVCE/1971; Humboldt, 1836/1967; Nietzsche, 1876/1983), this is often referenced as Sapir-Whorf hypothesis (SWH). Linguistic evidence in support of this hypothesis concentrated on conceptual contents of languages. For example, words for colors influence color perception (Roberson, Davidoff, & Braisbyb, 1999; Winawer,

Witthoft, Frank, Wu, Wade, & Boroditsky, 2007). The idea of language influencing cognition and culture has been criticized and "fell out of favor" in the 1960s (Wikipedia, 2009a) due to a prevalent influence of Chomsky's ideas emphasizing language and cognition to be separate abilities of the mind (Chomsky, 1965). Recently SWH again attracted academic attention, including experimental confirmations (see the previous references) and theoretical skepticism (Pinker, 2007). Interactions between language and cognition have been confirmed in fMRI experiments (Simmons, Stephan, Carla, Xiaoping, & Barsalou, 2008). Brain imaging experiments by Franklin, Drivonikou, Bevis, Davie, Kay, & Regier (2008) demonstrated that learning a word "rewires" cognitive circuits in the brain, learning a color name moves perception from right to left hemisphere. These recent data address in particular an old line of critique of SWH: whether the relationships between cultures and languages are causal or correlational; and if causal, what is the cause and what is the effect. Franklin et al (2008) experiments have demonstrated that language affects thinking, not the other way around. This discussion will be continued later but first I would like to emphasize that all arguments and experiments referenced above concentrated on conceptual effects of language.

Emotional effects might be no less important (Guttfreund, 1990; Harris, Ayçiçegi, & Gleason, 2003). In particular indicative are results of (Guttfreund, 1990): whereas the mother tongue is usually perceived as more emotional, Spanish-English bilinguals reported more intense emotions in psychological interviews conducted in Spanish than in English, irrespective of whether their first language was English or Spanish. Still, experimental evidence of interaction between the emotional contents of languages and cognition is limited, the neural mechanisms of these interactions are not known, and no computational models have existed (Perlovsky, 2006a,b,c; 2009a,b; 2010d,f,g,m).

This paper derives neurally motivated computational models of how conceptual and emotional contents of language affect cognition. The next section reviews conceptual and emotional mechanisms of language and its interaction with cognition. I briefly review existing theoretical ideas and experimental evidence on language evolution, conceptualizing possible mechanisms, and emphasizing directions for future research. Section 3 summarizes previously developed neuro-mathematical theories of interaction between language and cognition (Perlovsky, 2004; 2006a,b,c; 2009a,b; 2010d,f,g,m), which have been partially proven experimentally. These models are extended toward hierarchy of the mind. Section 4 derives neurally motivated cultural evolutionary models and demonstrates that different cultural evolutionary paths are favored by differences in interaction between cognition and language. In conclusion I discuss future theoretical and experimental directions.

## 2. Language and cognition

Language is widely considered as a mechanism for communicating conceptual information. Emotional contents of language are less appreciated and their role in the mind and evolutionary significance are less known. Still their roles in ontology, evolution, and cultural differences are significant. Whereas much research concentrates on language-computation, sensory-motor, and concept-intention interfaces (Hauser, Chomsky, & Fitch, 2002), this paper emphasizes that the primordial origin of language was a unified neural mechanism of fused voicing-behavior, emotion-motivation, and concept-understanding (Deacon, 1989; Lieberman, 2000; Mithen, 2007). It is likely that differentiation of mechanisms involved in language, voicing, cognition, motivation, and behavior occurred at different prehistoric times, in different lineages of our ancestors. This may be relevant to discussions of evolution of language and cognition (Botha, 2003; Botha & Knight 2009).

I address the current differentiated state of these abilities in the human mind, as well as unifying mechanisms of interfaces-links, which make possible integrated human functioning. The paper concentrates on mechanisms of existing interfaces and their cultural evolution. Before describing in the next section mechanisms of language, concepts, and emotions mathematically I will summarize these mechanisms conceptually in correspondence with general knowledge documented in a large number of publications emphasizing certain aspects that have escaped close scientific attention in the previous research.

2.1 Primordial undifferentiated synthesis

Animals' vocal tract muscles are controlled mostly from the ancient emotional center (Lieberman, 2000). Vocalizations are more affective than conceptual. Mithen (Mithen, 2007) summarized the state of knowledge about vocalization by apes and monkeys. Calls could be deliberate, however their emotional-behavioral meanings are probably not differentiated; primates cannot use vocalization separately from emotional-behavioral situations; this is one reason they cannot have language.

Emotionality of voice in primates and other animals is governed from a single ancient emotional center in the limbic system (Deacon, 1989; Lieberman, 2000; Mithen, 2007). Cognition is less differentiated than in humans. Sounds of animal cries engage the entire psyche, rather than concepts and emotions separately. An ape or bird seeing danger does not think about what to say to its fellows. A cry of danger is *inseparably* fused with recognition of a dangerous situation, and with a command to oneself and to the entire flock: "Fly!" An evaluation (emotion of fear), understanding (concept of danger), and behavior (cry and wing sweep) – are not differentiated. Conscious and unconscious are not separated. Recognizing danger, crying, and flying away is a fused concept-emotion-behavioral *synthetic* form of cognition-action. Birds and apes can not control their larynx muscles *voluntarily*.

2.2 Language and differentiation of emotion, voicing, cognition, and behavior

Origin of language required freeing vocalization from uncontrolled emotional influences. Initial undifferentiated unity of emotional, conceptual, and behavioral-(including voicing) mechanisms had to separate-differentiate into partially independent systems. Voicing separated from emotional control due to a separate emotional center in cortex which controls larynx muscles, and which is partially under volitional control (Deacon, 1989; Mithen, 2007). Evolution of this volitional emotional mechanism possibly paralleled evolution of language computational mechanisms. In contemporary languages the conceptual and emotional mechanisms are significantly differentiated, compared to animal vocalizations. The languages evolved toward conceptual contents, while their emotional contents were reduced. Cognition, or understanding of the world, is due to mechanisms of concepts, also referred to as mental representations or models. Barsalou calls this mechanism situated simulation (Barsalou, 2009). Perception or cognition consists of matching mental concept-models (simulations) with patterns in sensor data. Concept-models generate top-down neural signals that are matched to bottom-up signals coming from lower levels (Grossberg, 1988; Perlovsky, 2001) In this simulation process the vague mental models are modified to match concrete objects or situations (Perlovsky, 1987; 2001; 2006a; Bar, Kassam, Ghuman, Boshyan, Schmid, Dale, Hämäläinen, Marinkovic, Schacter, Rosen, and Halgren, 2006).

How these cognitive processes are determined and affected by language? Primate's cognitive abilities are independent from language. Language is fundamental to human cognitive abilities (Perlovsky, 2006a). A possible mathematical mechanism of language guiding and enhancing cognition have been discussed in (Perlovsky, 2004; 2006a,c; 2007a,b; 2009a,b; 2010d,g,m; Fontanari & Perlovsky, 2005; 2007; 2008a,b; Fontanari, Tikhanoff, Cangelosi, Perlovsky, & Ilin, 2009). This is a mechanism of the dual model whereby every concept-model has two parts: cognitive and language. The language models (words, phrases) are acquired from surrounding language by age of five or seven. They contain cultural wisdom accumulated through millennia. During the rest of life the language models guide the acquisition of cognitive models.

2.3 Emotions, instincts, and the knowledge instinct

The word emotion refers to several neural mechanisms in the brain (Juslin & Västfjäll, 2008); in this paper I always refer to instinctual-emotional mechanism described in (Grossberg & Levine, 1987). The word instinct in this paper is used in correspondence with this reference to denote a simple inborn, non-adaptive mechanism of internal "sensor," which measures vital body parameters, such as blood pressure, and indicates to the brain when these parameters are out of safe range. This simplified description will be sufficient for our purposes, more details could be found in (Grossberg & Seidman 2006; Gnadt & Grossberg, 2008) and references therein. We have dozens of such sensors, measuring sugar level in blood, body temperature, pressure at various parts, etc.

Mechanisms of concepts evolved for instinct satisfaction. According to instinctual-emotional theory (Grossberg & Levine, 1987), communicating satisfaction or dissatisfaction of instinctual needs from instinctual "sensors" to decision making parts of the brain is performed by emotional neural signals. Perception and understanding of concept-models corresponding to objects or situations that potentially can satisfy an instinctual need receive preferential attention and processing resources in the mind. In this paper emotions refer to neural signals connecting conceptual and instinctual brain regions.

Perception and cognition requires matching top-down signals from concept-models to bottom-up signals coming from sensory organs. Perception is required for survival. Therefore humans and high animals have an inborn drive to fit top-down and bottom-up signals, the knowledge instinct (KI; Perlovsky & McManus, 1991; Perlovsky, 1997; 2000; 2006a; 2007a). These references discuss specific emotions related to satisfaction or dissatisfaction of KI. These emotions are related purely to knowledge, not to bodily needs; since Kant (1790) this type of emotions are called aesthetic emotions. According to the theory of KI they are inseparable from every act of perception and cognition.

Biologists and psychologists have discussed various aspects of this mechanism: a need for positive stimulations, curiosity, a motive to reduce cognitive dissonance, a need for cognition (Harlow & Mears, 1979; Berlyne, 1960; Festinger, 1957; Cacioppo, Petty, Feinstein, & Jarvis, 1996; Levine & Perlovsky, 2008). Until recently, however, this drive was not mentioned among 'basic instincts' on a par with instincts for food and procreation. The fundamental nature of this mechanism became clear during mathematical modeling of workings of the mind. Our knowledge always has to be modified to fit the current situations. We don't usually see exactly the same objects as in the past: angles, illumination, and surrounding contexts are different. Therefore, our mental representations have to be modified; adaptation-learning is required (Grossberg, 1988; Kosslyn, Ganis, & Thompson, 2001).

All learning and adaptive algorithms maximize correspondence between the algorithm internal structure (knowledge in a wide sense) and objects of recognition; the psychological interpretation of this mechanism is KI. The mind-brain mechanisms of KI are discussed in (Levine & Perlovsky, 2008). As discussed below, it is a foundation of higher cognitive abilities, and it defines the evolution of consciousness and cultures.

2.4 Grammar, language emotionality, and meanings

Language and voice started separating from ancient emotional centers possibly millions of years ago. Nevertheless, emotions are present in language. Most of these emotions originate in cortex and are controllable aesthetic emotions. Their role in satisfying KI is considered in the next section. Emotional centers in cortex are neurally connected to old emotional limbic centers, so both influences are present. Emotionality of languages is carried in language sounds, what linguists call prosody or melody of speech. This ability of human voice to affect us emotionally is most pronounced in songs. Songs and music, however, is a separate topic (Perlovsky, 2006d; 2008; 2010a) not addressed in this paper.

Emotionality of everyday speech is low, unless affectivity is specifically intended. We may not notice emotionality of everyday "non-affective" speech. Nevertheless, "the right level" of emotionality is crucial for developing cognitive parts of models. If language parts of models were highly emotional, any discourse would immediately resort to fights and there would be no room for language development (as among primates). If language parts of models were non-emotional at all, there would be no motivational force to engage into conversations, to develop high cognitive models driven by language. Higher cognition would not be developed. Lower cognitive models, say for object perception would be developed because they are imperative for survival and because they can be developed independently from language, based on direct sensory perceptions, like in animals. But models of situations and higher cognition are developed based on language models (Perlovsky, 2004; 2006a,c; 2007b; 2009a; 2010f,g,m). As discussed later, this requires emotional connections between cognitive and language models.

Primordial fused language-cognition-emotional models, as discussed, have differentiated long ago. The involuntary connections between voice-emotion-cognition have dissolved with emergence of language. They have been replaced with habitual connections. Sounds of all languages have changed and, it seems, sound-emotion-meaning connections in languages should have severed. If the sounds of a language change slowly the connections between sounds and meanings persists and consequently the emotion-meaning connections persist. This persistence is a foundation of meanings because meanings imply motivations. If the sounds of a language change too fast, the cognitive models are severed from motivations, and meanings disappear. If the sounds change too slowly the meanings are nailed emotionally to the old ways, and culture stagnates.

This statement is a controversial issue, and indeed, it may sound puzzling. Doesn't culture direct language changes or is the language the driving force of cultural evolution? Direct experimental evidence is limited; it will have to be addressed by future research. Theoretical considerations suggest no neural or mathematical mechanism for culture directing evolution of language through generations; just the opposite, most of cultural contents are transmitted through language. Cognitive models contain cultural meanings separate from language (Perlovsky, 2009), but transmission of cognitive models from generation to generation is mostly facilitated by language. Cultural habits and visual arts can preserve and transfer meanings, but they contain a minor part of cultural wisdom and meanings comparative to those transmitted through the language. Language models are major containers of cultural knowledge shared among individual minds and collective culture.

The arguments in the previous two paragraphs suggest that an important step toward understanding cultural evolution is to identify mechanisms determining changes of the language sounds. As discussed below, changes in the language sounds are controlled by grammar. In inflectional languages, affixes, endings, and other inflectional devices are fused with sounds of word roots. Pronunciation-sounds of affixes are controlled by few rules, which persist over thousands of words. These few rules are manifest in every phrase. Therefore every child learns to pronounce them correctly. Positions of vocal tract and mouth muscles for pronunciation of affixes (etc.) are fixed throughout population and are conserved throughout generations. Correspondingly, pronunciation of whole words cannot vary too much, and language sound changes slowly. Inflections therefore play a role of "tail that wags the dog" as they anchor language sounds and preserve meanings. This, I think is what Humboldt (1836/1967) meant by "firmness" of inflectional languages. When inflections disappear, this anchor is no more and nothing prevents the sounds of language to become fluid and change with every generation.

This has happened with English language after transition from Middle English to Modern English (Lerer, 2007), most of inflections have disappeared and sound of the language started changing within each generation with this process continuing today. English evolved into a powerful tool of cognition unencumbered by excessive emotionality. English language spread democracy, science, and technology around the world. This has been made possible by conceptual differentiation empowered by language, which overtook emotional synthesis. But the loss of synthesis has also lead to ambiguity of meanings and values. Current English language cultures face internal crises, uncertainty about meanings and purposes. Many people cannot cope with diversity of life. Future research in psycholinguistics, anthropology, history, historical and comparative linguistics, and cultural studies will examine interactions between languages and cultures. Initial experimental evidence suggests emotional differences among languages consistent with our hypothesis (Guttfreund, 1990; Harris, Ayçiçegi, & Gleason, 2003).

Neural mechanisms of grammar, language sound, related emotions-motivations, and meanings hold a key to connecting neural mechanisms in the individual brains to evolution of cultures. Studying them experimentally is a challenge for future research. It is not even so much a challenge, because experimental methodologies are at hand; they just should be applied to these issues. The following sections develop mathematical models based on existing evidence that can guide this future research.

**3. Hierarchy of the mind and cultural dynamics**

This section summarizes mathematical models of the mind mechanisms corresponding to the discussion in the previous section. These models are based on the available experimental evidence and theoretical development by many authors summarized in (Perlovsky, 1987; 1994; 1997; 1998; 2000; 2006a,b,c; 2007b; 1009a,b; 1010d,f,g,m; Perlovsky, Plum, Franchi, Tichovolsky, Choi, & Weijers, 1997) and it corresponds to recent neuro-imaging data (Bar et al,

2006; Franklin et al, 2008).

Mechanisms of concepts, instincts, and emotions were described in above references and summarized in section 2.3. To briefly summarize, concepts operate like mental models of objects and situations; e.g., during visual perception of an object, a concept-model of the object stored in memory projects an image (top-down signals) onto the visual cortex, which is matched there to an image projected from retina (bottom-up signal). Perception occurs when top-down and bottom-up signals match. Concepts evolved for instinct satisfaction. The word instinct denotes here a simple inborn, non-adaptive mechanism of internal "sensor," which measures vital body parameters, such as blood pressure. Satisfaction or dissatisfaction of instinctual needs is communicated from instinctual parts of the brain to decision making parts of the brain by emotional neural signals. Perception and understanding of concept-models corresponding to objects or situations that potentially can satisfy an instinctual need receive preferential attention and processing resources. Mathematical description of these mechanisms were summarized in (Perlovsky, 2006a; 2007a).

Matching of top-down and bottom-up signals is described by dynamic logic; its fundamental aspect is the process "from vague to crisp." Mental representations are stored in memory as vague and distributed; they do not remind images of objects or situations. In the process of matching top-down and bottom-up signals, these vague representations are matched to sensory images and become crisp. This however is only possible at lower levels of concrete object perception. Object perception, in other words, is grounded in sensor data. At higher cognitive levels there is no concrete sensor data to ground cognition of abstract concepts. Higher level cognition is only possible due to language. Mental language representations are acquired from surrounding language, where they exist "ready-made." Learning of language does not require understanding of real life. For this reason language can be acquired early in life. This is why children learn language by 5 years of age, but cannot act like adults. The part of language instinct related to language learning is specific to humans; Pinker called it "the language instinct" (1994). The language instinct, however, does not connect language learning to real life. Cognitive representations connected to sensor and motor data are developed from life experience; the development of the hierarchy of these representations, far removed from direct sensor data, is only possible due to guidance by language. Cognitive representations connect language to life. This process (as any other mental process) could only move due to emotional motivations. Therefore, emotionality of language is crucial for connecting language to life (Perlovsky, 2006a; 2007a; 2009a,b; 2010d,f,g,m).

Let us emphasize this fundamental contradiction of the human mind mechanisms. Human cognition requires language. Evolution of language is only possible due to reduced emotionality of vocalization. Reduced emotionality constitutes the very possibility of language evolution and enables human cognition. However, if emotionality of a language becomes "too low," it is not connected to real life, and becomes void of meaning. Human thinking exists in this contradiction: abstract concepts that do not steer emotions are weakly connected to life experience.

## 4. Differentiation and synthesis

The fundamental contradiction of human mind described above affects the hierarchical dynamics of KI manifested as differentiation and synthesis. In interaction of top-down and bottom-up signals, at every layer of the hierarchy KI drives more crisp and detailed development of lower representations. At the same time KI drives higher representations toward more general and abstract ideas. Development of concrete and specific concepts is called differentiation, and creation of general concept-models, unifying differentiated signals is called synthesis.

Differentiation and synthesis are in complex relationships, at once symbiotic and antagonistic (Perlovsky, 2007a; 2009b). Synthesis creates emotional value of knowledge, it unifies language and cognition, and in this way it creates conditions for differentiation; it leads to spiritual inspiration, to active creative behavior leading to fast differentiation, creativity, knowledge, to science and technology. At the same time, a "too high" level of synthesis, "too" high emotional values of concepts stifles differentiation, as in traditional consciousness: every notion is so valuable emotionally that its differentiation becomes impossible.

Synthesis leads to growth of general concept-models and to growth of the hierarchy. This is counterbalanced by differentiation. Differentiation leads to the growth of the number of concepts approaching "precise knowledge about nothing". In the knowledge-acquiring regime the growth of synthesis is limited psychologically since the emotions of KI satisfaction "spread" over large number of concepts cannot sustain growing number of concepts, $D$. This is well known in many engineering problems, when too many models are used. Thus, whereas emotional synthesis creates a condition for differentiation (high emotional value of knowledge, efficient dual model connecting language and cognition, conceptual differentiation undermines synthesis (emotional value of knowledge). This interaction can be modeled by the following equations (Perlovsky 2009b):

$dD/dt = aD\ G(S),\ G(S) = (S - S_0)\ exp(-(S-S_0)/\ S_1),$
$dS/dt = -b\ D + d\ H,$
$H(t) = H_0 + e*t.$ (1)

Here, $t$ is time, $D$ is a number of concepts (differentiation), $S$ models synthesis, emotional satisfaction of KI, $H$ is a number of hierarchical levels; $a, b, d, e, S_0$ and $S_1$ are constants. Differentiation, $D$, grows proportionally to already existing number of concepts, as long as this growth is supported by synthesis, while synthesis is maintained at a "moderate" level, $S_0 < S < S_1$. "Too high" level of synthesis, $S > S_1$, stifles differentiation by creating too high emotional value of concepts. Synthesis, $S$, is related to emotion, but the detailed relationship will have to be established in future research. Synthesis, $S$, grows in the hierarchy, along with a number of hierarchical levels, $H$. By creating emotional values of knowledge, it sustains differentiation, however, differentiation, by spreading emotions among a large number of concepts destroys synthesis. Detailed model of hierarchical dynamics $H$ is difficult, so instead we just consider a period of slow growth of the hierarchy $H$. At moderate values of synthesis, solving eqs.(1) yields a solution in Fig. 1a. The number of concepts grows until certain level, when it results in reduction of synthesis; then the number of models falls. As a number of models falls, synthesis grows, and the growth in models resumes. The process continues with slowly growing, oscillating number of models. Oscillations affecting up to 80% of knowledge indicate internal instability of knowledge-accumulating consciousness. Significant effort was extended to find solutions with reduced oscillations, however, no stable knowledge-acquiring solution was found based on eqs.(1). This discussion is continued below (Fig.2).

Another solution corresponds to initially high level of synthesis, Fig. 1b. Synthesis continues growing whereas differentiation levels off. This leads to a more and more stable society with high synthesis, in which high emotional values are attached to every concept, however, differentiation stagnates.

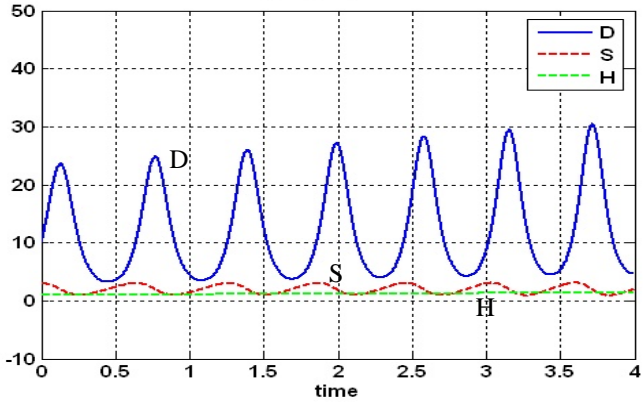 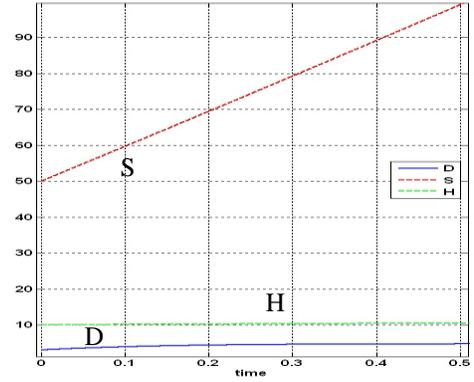

Fig.1a. Conceptual-differentiating culture     Fig.1b. Emotional-traditional culture

Fig. 1. (a) Evolution of culture at moderate values of synthesis oscillates: periods of flourishing and knowledge accumulation alternate with collapse and loss of knowledge ($a = 10$, $b = 1$, $d = 10$, $e = 0.1$, $S_0=2$, $S_1=10$, and initial values $D(t=0) = 10$, $S(t=0) = 3$, $H_0 = 1$; parameter and time units are arbitrary). In long time the number of models slowly accumulates; this corresponds to slowly growing hierarchy. (b) Evolution of highly stable, stagnating society with growing synthesis. High emotional values are attached to every concept, while knowledge accumulation stops ($M(t=0)= 3$, $H_0 = 10$, $S(t=0) = 50$, $S_0 = 1$, $S_1 = 10$, $a = 10$, $b = 1$, $d = 10$, $e=1$).

These two solutions of eqs.(1) can be compared to Humboldt's (1836/1967) characterization of languages and cultures. He contrasted inert objectified "outer form" of words vs. subjective, culturally conditioned, and creative "inner form." Humboldt's suggestion continues to stir linguists' interest today, yet seem mysterious and not understood scientifically.

This paper analysis suggests the following interpretation of Humboldt's thoughts in terms of neural mechanisms. His "inner form" corresponds to the integrated, moderately emotional neural dual model (Perlovsky, 2004; 2006c; 2007b; 2009a,b; 2010g). Contents of cognitive models are being developed guided by language models, which accumulate cultural wisdom, Fig. 1a. "Outer form" of language corresponds to inefficient state of neural dual model, in which language models do not guide differentiation of the cognitive ones. This might be due to either too strong or too weak involvement of emotions. If emotional involvement in cognition or language is too weak, learning does not take place because motivation disappears. If emotional involvement is too strong, learning does not take place because old knowledge is perceived as too valuable, and no change is possible. The first case might be characteristic of low-inflected languages, when sound of language changes "too fast," and emotional links between sound and meanings are severed. The second case might be characteristic of "too strongly" inflected languages, in which sound changes "too slowly" and emotions are connected to meanings "too strongly;" this could be a case of Fig. 1b. A brief look at cultures and languages certainly points to many examples of this case: highly inflected languages and correspondingly "traditional" stagnating cultures. Which of these correspond to Fig. 1b and the implied neural mechanisms? What it means quantitatively: "too fast" or "too slow," and which cultures and languages correspond to which case will require further psycholinguistic and anthropological research.

Preliminary analysis indicates English as a typical language corresponding to Fig. 1a. and Arabic corresponding to Fig. 1b. This corresponds to Humboldtian analysis. The integrated dual model assumes "moderate" emotional connection between language and cognitive models, which fosters the integration and does not impede it. Humboldt suggested that this relationship is characteristic of inflectional languages (such as Indo-European), inflection provided "the true inner firmness for the word with regard to the intellect and the ear" (today we would say "concepts and emotions"). The integrated dual model assumes a moderate value of synthesis, Fig. 1a, leading to interaction between language and cognition and to accumulation of knowledge. This accumulation, however, does not proceed smoothly; it leads to instabilities and oscillations, possibly to cultural calamities; this characterizes significant part of European history from the fall of Roman Empire to recent times.

Much of contemporary world is "too flat" for an assumption of a single language and culture, existing without outside influences. Fig. 2 demonstrates an evolutionary scenario for two interacting cultures that exchange differentiation and synthesis; for this case eqs. (1) are modified by adding $xD$ to the first equation and $yS$ to the second, where $x$ and $y$ are small constants, while $D$ and $S$ are taken from the other culture. The first and second cultures initially corresponded to Figs.1a and 1b correspondingly. After the first period when the influence of the first culture dominated, both cultures stabilized each other, both benefited from fast growth and reduced instabilities.

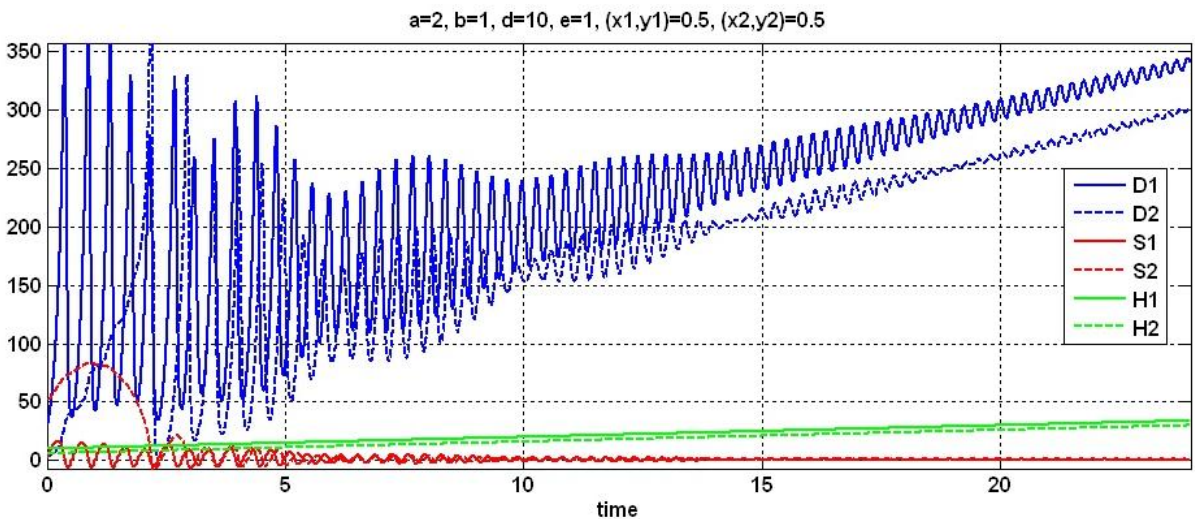

Fig. 2. Effects of cultural exchange ($k=1$, solid lines: $M(t=0)= 30$, $H0 = 12$, $S(t=0) = 2$, $S0 = 1$, $S1 = 10$, $a = 2$, $b = 1$, $d = 10$, $e=1$, $x = 0.5$, $y = 0.5$; $k=2$, dotted lines: $M(t=0)= 3$, $H0 = 10$, $S(t=0) = 50$, $S0 = 1$, $S1 = 10$, $a = 2$, $b = 1$, $d = 10$, $e=1$, $x = 0.5$, $y = 0.5$). Transfer of differentiated knowledge to less-differentiated culture dominates exchange during $t < 2$ (dashed blue curve). In long run $(t > 5)$, cultures stabilize each other, and swings of differentiation and synthesis subside while knowledge accumulation continues.

## 5. Discussion and future research

Connections between the neural mechanisms, language, emotions, and cultural evolution proposed in this paper are but a first step requiring much experimental evidence and theoretical development. Influence of language on culture, the Bhartrihari-Humboldt-Nietzsche-Sapir-Whorf hypothesis formalized by the discussed mechanism adds a novel aspect to this old idea.

The emotional contents of languages could be more important in influence on cultures than their conceptual contents.

In the milieu defined by Chomsky's assumed independence of language and cognition the Sapir-Whorf hypothesis (SWH) has steered much controversy:

"This idea challenges the possibility of perfectly representing the world with language, because it implies that the mechanisms of *any* language condition the thoughts of its speaker community" (Wikipedia, 2008).

The fact that Wikipedia seriously considers a naïve view of "perfectly representing the world" as a scientific possibility is indicative of a problematic state of affairs, "the prevalent commitment to uniformitarianism, the idea that earlier stages of languages were just as complex as modern languages" (Hurford, 2008). With the development of cognitive and evolutionary linguistics diversity of languages are considered in their evolutionary reality, and identifying neural mechanisms of language evolution and language-cognition interaction is coming in demand.

Neural mechanisms proposed in this paper and models inspired by these mechanisms are but an initial step in this line of research. Nevertheless concrete predictions are made for relations between language grammars and types of cultures. These predictions can be verified in psycholinguistic laboratories, eq.(1) coefficients can be measured using existing methods.

Future mathematical-theoretical research should address continuing development of both mean-field and multi-agent simulations, connecting neural and cultural mechanisms of emotions and cognition and their evolution mediated by language. KI theory should be developed toward theoretical understanding of its differentiated forms explaining multiplicity of aesthetic emotions in language prosody and music (Perlovsky, 2006d; 2008; 2010a). This theoretical development should go along with experimental research clarifying neural mechanisms of KI (Levine & Perlovsky, 2008; Bar et al, 2006) and the dual language-cognitive model, (Perlovsky, 2009).

Recent experimental results on neural interaction between language and cognition (Franklin et al, 2008; Simmons et al, 2008) support the mechanism of the dual model. Brain imaging can be used to directly verify the dual model. These methods should be expanded to interaction of language with emotional-motivational, voicing, behavioral, and cognitive systems

Prehistoric anthropology should evaluate the proposed hypothesis that the primordial system of fused conceptual cognition, emotional evaluation, voicing, motivation, and behavior differentiated at different prehistoric time periods. Are there data to support this hypothesis, can various stages of prehistoric cultures be associated with various neural differentiation stages? Can different humanoid lineages be associated with different stages of neural system differentiation? What stage of neural differentiation corresponds to Mithen's hypothesis about singing Neanderthals (Mithen, 2007)? Psychological social and anthropologic research should go in parallel documenting various cultural evolutionary paths and correlations between cognitive and emotional contents of historical and contemporary cultures and languages.

Proposed correlation between grammar and emotionality of languages can be verified in direct experimental measurements using skin conductance and fMRI neuro-imaging. Emotional version of Sapir-Whorf hypothesis should be evaluated in parallel psychological and anthropological research. More research is needed to document cultures stagnating due to "too" emotional languages; as well as crises of lost values due to "low" emotionality of languages.

**Acknowledgment**

I am thankful to M. Alexander, M. Bar, R. Brockett, M. Cabanac, R. Deming, F. Lin, J. Gleason, R. Kozma, D. Levine, A. Ovsich, and B. Weijers, and to AFOSR PMs Dr. Jon Sjogren and Dr. Doug Cochran for supporting part of this research.